%% file: main.tex
\documentclass[conference]{IEEEtran}
\IEEEoverridecommandlockouts
\usepackage{cite}
\usepackage{amsmath,amssymb,amsfonts}
\usepackage{algorithmic}
\usepackage{graphicx}
\usepackage{textcomp}
\usepackage{xcolor}
\usepackage[caption=true, font=footnotesize]{subfig}
\usepackage{balance}
\usepackage{multirow}
\def\BibTeX{{\rm B\kern-.05em{\sc i\kern-.025em b}\kern-.08em
    T\kern-.1667em\lower.7ex\hbox{E}\kern-.125emX}}
\begin{document}

\title{\textcolor{black}{Using Intuition from} Empirical Properties to Simplify Adversarial Training Defense}

\author{\IEEEauthorblockN{Guanxiong Liu}
\IEEEauthorblockA{\textit{ECE Department} \\
\textit{New Jersey Institute of Technology}\\
Newark, USA \\
gl236@njit.edu}
\and
\IEEEauthorblockN{Issa Khalil}
\IEEEauthorblockA{\textit{QCRI} \\
\textit{Hamad bin Khalifa University}\\
Doha, Qatar \\
ikhalil@hbku.edu.qa}
\and
\IEEEauthorblockN{Abdallah Khreishah}
\IEEEauthorblockA{\textit{ECE Department} \\
\textit{New Jersey Institute of Technology}\\
Newark, USA \\
abdallah@njit.edu}
}
\maketitle

\input{abstract}

\input{introduction}

\input{property-1}

\input{property-2}

\input{methodology}

\input{results}

\input{conclusion}

\bibliographystyle{IEEEtranS}
\bibliography{reference}

\end{document}

%% file: abstract.tex
\begin{abstract}\label{sec:abstract}
Due to the surprisingly good representation power of complex distributions, neural network (NN) classifiers are widely used in many tasks which include natural language processing, computer vision and cyber security. In recent works, people noticed the existence of adversarial examples. These adversarial examples break the NN classifiers' underlying assumption that the environment is attack free and can easily mislead fully trained NN classifier without noticeable changes. Among defensive methods, adversarial training is a popular choice. However, original adversarial training with single-step adversarial examples (Single-Adv) can not defend against iterative adversarial examples. Although adversarial training with iterative adversarial examples (Iter-Adv) can defend against iterative adversarial examples, it consumes too much computational power and hence is not scalable. In this paper, we analyze Iter-Adv techniques and identify two of their empirical properties. Based on these properties, we propose modifications which enhance Single-Adv to perform competitively as Iter-Adv. Through preliminary evaluation, we show that the proposed method enhances the test accuracy of state-of-the-art (SOTA) Single-Adv defensive method against iterative adversarial examples by up to 16.93\% while reducing its training cost by 28.75\%.
\end{abstract}


%% file: introduction.tex
\section{Introduction}\label{sec:introduction}

Adversarial examples were discovered by Szegedy et. al. and presented in \cite{szegedy2013intriguing}. In the image classification tasks, they show that a specially designed perturbation which can be ignored by human eyes can effectively mislead the fully trained NN classifier. Moreover, such perturbation is not a special case but can be found for almost every example. Yet, more scary, the research shows that adversary could arbitrarily control the prediction from NN classifier through carefully designed perturbations and can achieve high success rate against classifiers without defenses \cite{carlini2016towards}\cite{kurakin2016adversarial1}\cite{madry2017towards}.

Thereafter, great effort has been devoted to designing an effective method to defend against adversarial examples. Some of these methods utilize augmentation and regularization to enhance test accuracy on adversarial examples \cite{papernot2016distillation}. Other methods rely on building a protective shells around the classifier to either identify adversarial examples or mitigate the adversarial perturbations \cite{meng2017magnet}\cite{samangouei2018defense}. Among all existing defensive approaches, adversarial training is shown to be the most successful because unlike many other defensive approaches it does not rely on the false sense of security brought by obfuscated gradient \cite{athalye2018obfuscated}. The fundamental idea is using adversarial examples during the training. The stronger the adversarial examples used the stronger the obtained defense. Therefore, the research community originally performs adversarial training with single-step adversarial examples (Single-Adv) and now with iterative adversarial examples (Iter-Adv) \cite{goodfellow2014explaining}\cite{kurakin2016adversarial2}\cite{madry2017towards}.

  One of the biggest problems of Iter-Adv is the huge computational cost in preparing iterative adversarial examples during the training \cite{athalye2018obfuscated}. For example, using Iter-Adv on ImageNet dataset requires a cluster of GPU servers \cite{kannan2018adversarial}. Nowadays, there is a trend to make the NN classifier more portable such that it can be trained and utilized solely in a smart-phone \cite{chen2015mxnet}. Moreover, due to data privacy consideration, some applications calculate the last few layers of NN on local device. Under these scenarios, we need a lightweight adversarial training defense since Iter-Adv is hard to be scaled with limited computational resources \cite{athalye2018obfuscated}. In the following sections, we start with two questions about Iter-Adv. Then, based on our empirical results, we propose two modifications to Iter-Adv and share our idea to simplify it. Our contributions are summarized as follows: 

\begin{itemize}
    \item We raise two questions about Iter-Adv and conduct experiments that identify two empirical properties of it.
    
    \item Based on identified properties, we propose a Single-Adv method that performs competitively as Iter-Adv methods.
    
    \item Through comparison with SOTA Single-Adv method, our proposed method enhances test accuracy by 16.93\% and reduces training time by 28.75\%.
\end{itemize}

The rest of the work is organized as follows. Section \ref{sec:property-1} and \ref{sec:property-2} present our questions and the preliminary experiments. Section \ref{sec:methodology} proposes modifications that simplify Iter-Adv. Section \ref{sec:results} presents our preliminary evaluation results. Section \ref{sec:conclusion} and \ref{sec:future} conclude the paper and present the future work.

%% file: property-1.tex
\begin{figure*}[ht]
\centering
\begin{minipage}[c]{.32\textwidth}
    \begin{minipage}[c]{\textwidth}
    \centering
        \subfloat[\label{fig:q1-mnist}]{
            \includegraphics[width=\linewidth]{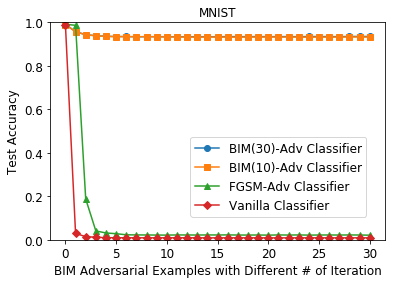}}
    \end{minipage}
    \begin{minipage}[c]{\textwidth}
    \centering
        \subfloat[\label{fig:q1-fmnist}]{
            \includegraphics[width=\linewidth]{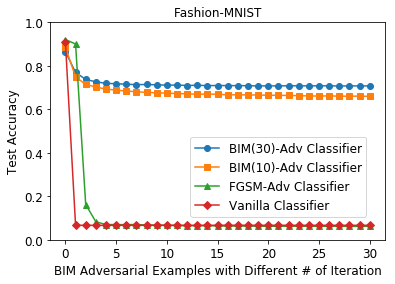}}
    \end{minipage}
\captionsetup{width=0.7\textwidth}
\caption{Test Accuracy on BIM Examples with Different Numbers of Iteration}
\label{fig:q1-res}
\end{minipage}
\begin{minipage}[c]{.32\textwidth}
    \begin{minipage}[c]{\textwidth}
    \centering
        \subfloat[\label{fig:q2-mnist}]{
            \includegraphics[width=\linewidth]{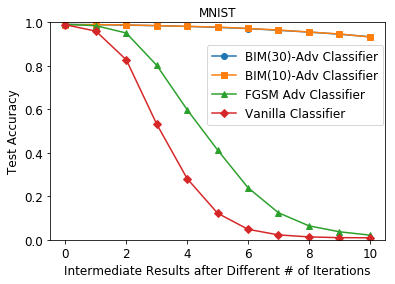}}
    \end{minipage}
    \begin{minipage}[c]{\textwidth}
    \centering
        \subfloat[\label{fig:q2-fmnist}]{
            \includegraphics[width=\linewidth]{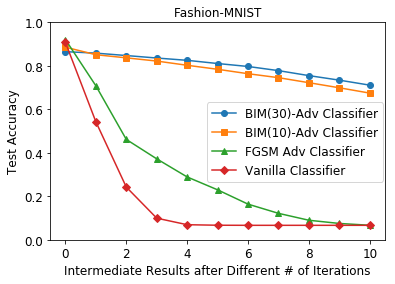}}
    \end{minipage}
\captionsetup{width=0.7\textwidth}
\caption{Test Accuracy on Intermediate BIM Examples after Each Iteration}
\label{fig:q2-res}
\end{minipage}
\begin{minipage}[c]{.32\textwidth}
    \begin{minipage}[c]{\textwidth}
    \centering
        \subfloat[\label{fig:adv-flow}]{
            \includegraphics[width=\linewidth]{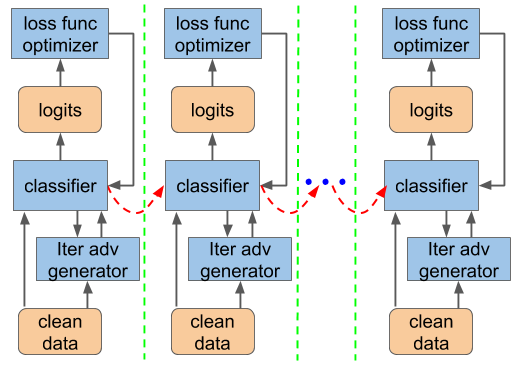}}
    \end{minipage}
    \begin{minipage}[c]{\textwidth}
    \centering
        \subfloat[\label{fig:diat-flow}]{
            \includegraphics[width=\linewidth]{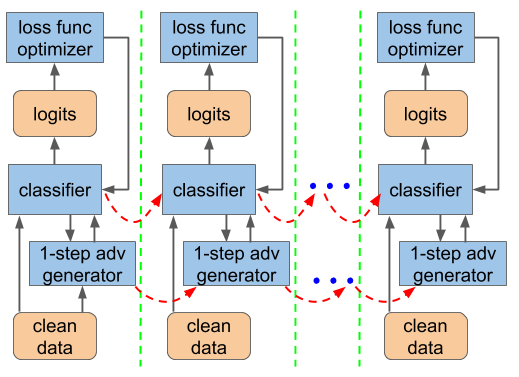}}
    \end{minipage}
\captionsetup{width=0.7\textwidth}
\caption{Flow Chart of Iter-Adv and the Proposed Method}
\label{fig:flow-chart}
\end{minipage}
\vspace{-5mm}
\end{figure*}

\section{Is it worth to use tiny per step perturbations?}\label{sec:property-1}
Let's take a step back and recall the definition of gradient based $l_{\infty}$ iterative adversarial examples. 
\begin{gather*}
    \delta_{i} = sign(\nabla_{x_{i-1}} \mathcal{L}(\mathcal{C}(x_{i-1}, \theta), y)) \times \epsilon_{i}\\
    x_{i} = clip(x_{i-1} + \delta_{i})
\end{gather*}
where $x_{0}$ is the original example, $x_{i}$ is the $i$th iteration adversarial example, $y$ is the ground truth, $\mathcal{C}$ is the classifier, $\mathcal{L}$ is the loss function, $\epsilon_{i}$ is the perturbation limit in the $i$th iteration, and $\delta_{i}$ is the calculated perturbation in the $i$th iteration.

To generate iterative adversarial examples, adversaries apply small per step perturbation several times and update the gradient direction based on their observation of the targeted NN after each step. Generally speaking, the smaller per step perturbation they apply the better observation of NN's decision hyperplane they may get. With a better observation of NN's decision hyperplane, adversaries could greedily optimize their objective function and generate more serious adversarial examples.

However, there are still several questions to be answered. \textit{Given the relation between iterative adversarial examples and the per step perturbation, how much can we benefit from a smaller per step perturbation? Is there a limit on the per step perturbation beyond which the improvement starts to vanish?}

To answer these questions, we conduct experiments on MNIST and Fashion-MNIST datasets, respectively. On each dataset, we firstly train four different NN classifiers with the same structure and hyper-parameter setting \cite{song2018improving}. These include: (1) a Vanilla classifier trained on original examples only, (2) a FGSM-Adv classifier trained on a mixture original and FGSM examples \cite{goodfellow2014explaining}, (3) two BIM($\cdot$)-Adv classifiers trained on a mixture of original and BIM (different numbers of iteration) examples \cite{kurakin2016adversarial2}. The number of iterations for the two BIM-Adv classifiers is $10$ and $30$, respectively. We evaluate these classifiers in terms of test accuracy when facing BIM examples with different numbers of iteration, $N$. The total perturbation, $\epsilon$, is fixed to $0.3$ (MNIST) and $0.2$ (Fashion-MNIST) with $l_{\infty}$ norm. The per step perturbation is set to $\epsilon_{s} = \frac{\epsilon}{N}$. The evaluation results are presented in Figure \ref{fig:q1-res}.

From Figure \ref{fig:q1-res}, it is clear that test accuracy results of all four classifiers converge quite fast. The Vanilla and FGSM-Adv classifiers can not defend BIM examples and their test accuracy results drop below $10\%$ (random guessing) when the iteration number is larger than $4$ on both MNIST and Fashion-MNIST datasets. The BIM-Adv classifiers are shown to be defensive against BIM examples and their test accuracy results also converge after we set iteration number to $5$ (MNIST) and $10$ (Fashion-MNIST). Given the fact that adversarial training uses adversarial examples to find blind spots of the under-training classifier and train it, these experiments show that \textbf{(1) there is a limit for decreasing the per step perturbation of iterative adversarial examples and (2) iterative adversarial examples with per step perturbation lower than the limit only marginally benefit the adversarial training.} This conclusion can also be verified through comparing test accuracy results of BIM-Adv classifiers. Although BIM($30$)-Adv is trained on BIM examples with much smaller per step perturbation compared with BIM($10$)-Adv, they both have almost the same test accuracy on MNIST dataset. On Fashion-MNIST, BIM($30$)-Adv is shown to be more defensive than BIM($10$)-Adv. However, the difference on test accuracy does not increase with the iteration number. The reason is that BIM($10$)-Adv has a stable test accuracy even when facing BIM examples with smaller per step perturbation ($N>10$).

Based on the results and the conclusion in this subsection, we proposed the following. When facing the trade-off between defensive performance and computational efficiency in training with iterative adversarial examples, the per step perturbation is not necessary to be very small since it only marginally benefits the adversarial training.

%% file: property-2.tex
\section{Any benefit of using intermediate results?}\label{sec:property-2}
Adversarial training originally uses single-step adversarial examples and recently shifts to iterative adversarial examples. Since iterative adversarial examples are more serious and able to reveal more blind spots of under-training classifier, such shift improves the adversarial training in building classifier's defense against iterative adversarial examples.

However, there are still several questions to be answered during this shift. \textit{When we prepare iterative adversarial examples for training, can we also benefit from using intermediate results from this generation process as well?}

To explore the answer for this question, we conduct another set of experiments on MNIST and Fashion-MNIST datasets, respectively. On each dataset, we continuously use the same NN classifiers (one Vanilla classifier, one FGSM-Adv classifier and two BIM-Adv classifiers) as before. During the evaluation, we test all four classifiers in terms of test accuracy against BIM examples. The total perturbation of test examples is set to $\epsilon=0.3$ (MNIST) and $\epsilon=0.2$ (Fashion-MNIST). Different from previous experiments, we generate BIM examples with fixed iteration number, $N=10$, and evaluate the test accuracy after each iteration. Therefore, the per step perturbation is fixed to $\epsilon_{s} = \frac{\epsilon}{10}$ while perturbation is increasing after each iteration. The evaluation results are presented in Figure \ref{fig:q2-res}.

The experiment results from Figure \ref{fig:q2-res} show that the test accuracy of all four classifiers is monotonically decreasing with the number of iterations. The classifiers without defense, Vanilla and FGSM-Adv, are defeated (perform worse than random guessing) by the intermediate results before the iterative adversarial examples are ready (around $8$ iterations on both MNIST and Fashion-MNIST datasets). Although two BIM-Adv classifiers obtain defense from training and can correctly classify most of adversarial examples, the majority of test accuracy degeneration still happens within first couple of iterations (about $6$ iterations on both MNIST and Fashion-MNIST datasets). Given the fact that adversarial training uses adversarial examples to find blind spots of the under-training classifier and train it, these experiment results show that \textbf{(1) the majority of blind spots can be revealed by intermediate results during generating iterative adversarial examples and (2) using intermediate results can benefit adversarial training before iterative adversarial examples are ready.}

Based on the results and the conclusion they leads to, we can summarize our second suggestion as follows. To enhance the training efficiency in Iter-Adv, we could utilize the intermediate results in training before the iterative adversarial examples are ready since the majority of weaknesses can be revealed by them at that time. It is worth to clarify that we do not suggest to fully replace iterative adversarial examples with intermediate results but the utilization of intermediate results can reduce the total computation in Iter-Adv.

%% file: methodology.tex
\section{Simplifying Adversarial Training Defense}\label{sec:methodology}

In previous sections, we raise two questions regarding the training with iterative adversarial examples and provide intuitions to answer them based on the results from the MNIST and the Fashion-MNIST datasets. In this section, we propose our modifications to simplify Iter-Adv which takes both defensive performance and computational efficiency into consideration.

Before moving further, we want to firstly review the working process of Iter-Adv. As we can see from Figure \ref{fig:adv-flow}, a flow chart of adversarial training is provided. At the beginning, the clean examples are fed into the classifier and the generator of adversarial examples. The generator then interacts with the classifier for several iterations to prepare iterative adversarial examples. When adversarial examples are ready, the classifier generates prediction logits for both clean and adversarial examples through forward propagation \cite{goodfellow2016deep}. Finally, a predefined loss function optimizer takes these prediction logits to calculate the loss value and update NN parameters through gradient descent and backward propagation \cite{goodfellow2016deep}. During the training, these steps are repeated for several epochs (separated by the green dash line) and NN parameters in classifier are carried through epochs (the red dash line with arrow).

Inspired by the summarized empirical properties, we now propose two modifications to simplify Iter-Adv. The flow chart of our proposed method can be found in Figure \ref{fig:diat-flow}. Compared with Iter-Adv, our first modification is utilizing the intermediate results during the training. In each training epoch, our proposed method will not wait until the final iterative adversarial examples are ready. Instead, it utilizes the intermediate results and carry them to the next training epoch. As a result, it only requires a single-step perturbation in each epoch. The second modification is selecting a relatively large per step perturbation. With this modification, the adversarial examples can quickly reach large perturbation and reveal the majority of blind spots. Therefore, it mitigates the disadvantage brought by using single-step adversarial perturbation, training with weak adversarial examples in the first few training epochs. To catch up the long term changes in classifier's parameters, this epoch-wise iteration process will be reset after a certain number of training epochs.

From Figure \ref{fig:adv-flow} and Figure \ref{fig:diat-flow}, it is clear that our proposed method significantly reduces computational cost required by preparing iterative adversarial examples in each training epoch. By reusing the adversarial examples in the next training epoch and selecting large per step perturbation, we can also expect the proposed method to perform better than FGSM-Adv and closer to BIM-Adv.

%% file: results.tex

%
\begin{table*}[t!]
    \begin{center}
    \begin{tabular}{ c | c  c  c  c | c  c  c  c | c }
    \hline \hline
    & \multicolumn{4}{c |}{MNIST} & \multicolumn{4}{c |}{Fashion-MNIST} & \multirow{2}{0.12\linewidth}{Training Time per Epoch (seconds)}\\
    & Original & FGSM & BIM(10) & BIM(30) & Original & FGSM & BIM(10) & BIM(30) &  \\
    \hline
    FGSM-Adv & 99.10\% & 98.65\% & 2.16\% & 2.13\% & 92.09\% & 90.05\% & 6.68\% & 6.18\% & 17.12 \\
    ATDA & 97.80\% & 97.90\% & 86.6\% & 84.2\% & 85.60\% & 80.3\% & 56.8\% & 49.1\% & 26.21 \\
    Proposed & 99.08\% & 96.71\% & 94.21\% & 94.04\% & 89.19\% & 79.85\% & 70.74\% & 66.03\% & 18.68 \\
    \hline
    BIM(10)-Adv & 98.97\% & 96.74\% & 93.29\% & 93.24\% & 88.59\% & 74.93\% & 67.45\% & 65.96\% & 56.47 \\
    BIM(30)-Adv & 99.06\% & 95.71\% & 93.39\% & 93.51\% & 86.52\% & 76.98\% & 71.12\% & 70.75\% & 142.37 \\
    \hline \hline
    \end{tabular}
    \end{center}
    \caption{Evaluation Results}
    \label{table:summary-test-accuracy}
    \vspace{-5mm}
\end{table*}

\section{Preliminary Results}\label{sec:results}

In this section, we evaluate the proposed method. Since our proposed method belongs to Single-Adv methods, we compare it with several Single-Adv and Iter-Adv methods. Iter-Adv methods include FGSM-Adv, BIM(10)-Adv and BIM(30)-Adv. We also include a SOTA Single-Adv method which is called ATDA \cite{song2018improving}. This method trains the classifier with single-step adversarial examples and modifies training loss to achieve domain adaptation. Our proposed method trains the classifier with single-step adversarial examples and the perturbation is limited at $\frac{\epsilon}{10}$. Moreover, the epoch-wise iteration presented in Figure \ref{fig:diat-flow} is repeated every $20$ training epochs. The results from Single-Adv methods are represented by the blue color bars while the results from Iter-Adv methods are represented by the green color bars.

The evaluation is conducted on MNIST and Fashion-MNIST datasets and measures both defensive power and computational consumption in training. To measure defensive power, we evaluate different methods against two white-box $l_{\infty}$ iterative adversarial examples which are utilized in Section \ref{sec:property-1}, BIM(10) and BIM(30). Since all methods are run on the same workstation with a Tesla K20m GPU and converge after the same number of epochs, we utilize the training time per epoch as a measure of computational consumption.

From the results presented in Table \ref{table:summary-test-accuracy}, it is clear that all methods can correctly classify original and FGSM adversarial examples while only ATDA, BIM(10)-Adv, BIM(30)-Adv and our proposed method show resistance against iterative adversarial examples. Our proposed method constantly outperforms ATDA in term of test accuracy. Under MNIST dataset, the enhancements are 7.61\% on BIM(10) and 9.83\% on BIM(30). Compared with BIM(10)-Adv and BIM(30)-Adv, our proposed method achieves the same level and even slightly higher test accuracy. When the experiments are extended to Fashion-MNIST dataset, the advantage of our proposed method over ATDA is even more significant. On BIM(10) and BIM(30) adversarial examples, respectively, our proposed method enhances the test accuracy by 13.94\% and 16.93\% compared to ATDA. More importantly, the performance of our proposed method is still competitive compared with BIM(10)-Adv and BIM(30)-Adv in terms of test accuracy. 

Beside the test accuracy, we also evaluate all defensive methods on average training time per epoch. From the results presented in Table \ref{table:summary-test-accuracy}, it is clear that Single-Adv methods significantly reduce the training time consumed by Iter-Adv by up to 85\%. More importantly, our proposed method can further reduce 28.75\% of that required by the SOTA Single-Adv method, ATDA.

%% file: conclusion.tex
\section{Conclusion}\label{sec:conclusion}

In this work, we raise two questions about Iter-Adv and identify two empirical properties. (1) Iterative adversarial examples with per step perturbation lower than a certain limit only marginally benefit the adversarial training. (2) The majority of blind spots can be revealed by intermediate results during generating iterative adversarial examples. Based on these two properties, we propose a Single-Adv method which performs competitively as Iter-Adv methods. Through preliminary results, the proposed method outperforms the SOTA Single-Adv method (by up-to 16.93\% in test accuracy) and achieves the same level of defense as Iter-Adv methods. More importantly, the proposed method significantly reduces the training time required by Iter-Adv methods and even save 28.75\% of that required by the SOTA Single-Adv method.

\section{Future Work}\label{sec:future}

In the future, we are going to extend the experiments on larger datasets and add more experiments to get deeper understanding of Single-Adv and Iter-Adv.